\def\year{2021}\relax
\documentclass[letterpaper]{article} % DO NOT CHANGE THIS
\usepackage{aaai21}  % DO NOT CHANGE THIS
\usepackage{times}  % DO NOT CHANGE THIS
\usepackage{helvet} % DO NOT CHANGE THIS
\usepackage{courier}  % DO NOT CHANGE THIS
\usepackage[hyphens]{url}  % DO NOT CHANGE THIS
\usepackage{graphicx} % DO NOT CHANGE THIS
\usepackage{natbib}  % DO NOT CHANGE THIS AND DO NOT ADD ANY OPTIONS TO IT
\usepackage{caption} % DO NOT CHANGE THIS AND DO NOT ADD ANY OPTIONS TO IT
\usepackage{booktabs}       % professional-quality tables
\usepackage{amsfonts}       % blackboard math symbols
\usepackage{amsmath}
\usepackage{bm}
\usepackage{wrapfig}  % <- we shoul remove this package

\title{Query Training: Learning a Worse Model to Infer Better Marginals\\ in Undirected Graphical Models with Hidden Variables}

\author {
    Miguel Lázaro-Gredilla, \textsuperscript{\rm 1}
    Wolfgang Lehrach, \textsuperscript{\rm 1}
    Nishad Gothoskar, \textsuperscript{\rm 2}\\
    Guangyao Zhou, \textsuperscript{\rm 1}
    Antoine Dedieu, \textsuperscript{\rm 1}
    Dileep George\textsuperscript{\rm 1}\\
}
\affiliations {
    \textsuperscript{\rm 1} Vicarious AI \\
    \textsuperscript{\rm 2} MIT \\
    \{miguel, wolfgang, stannis, antoine, dileep\}@vicarious.com, nishadg@mit.edu
}

\begin{document}

\maketitle

\begin{abstract}
Probabilistic graphical models (PGMs) provide a compact representation of knowledge that can be queried in a flexible way: after learning the parameters of a graphical model once, new probabilistic queries can be answered at test time without retraining. However, when using undirected PGMS with hidden variables, two sources of error typically compound in all but the simplest models (a) learning error (both computing the partition function and integrating out the hidden variables is intractable); and (b) prediction error (exact inference is also intractable). Here we introduce query training (QT), a mechanism to learn a PGM that is optimized for the approximate inference algorithm that will be paired with it. The resulting PGM is a worse model of the data (as measured by the likelihood), but it is tuned to produce better marginals for a given inference algorithm. Unlike prior works, our approach preserves the querying flexibility of the original PGM: at test time, we can estimate the marginal of any variable given any partial evidence. We demonstrate experimentally that QT can be used to learn a challenging 8-connected grid Markov random field with hidden variables and that it consistently outperforms the state-of-the-art AdVIL when tested on three undirected models across multiple datasets.
\end{abstract}

\section{Introduction}
\label{sec:intro}

Probabilistic graphical models (PGMs) define probability density functions (PDFs) over data. Their graph structure describes the conditional dependencies between the random variables of the model, and how the joint PDF can be expressed as a product of factors, with each factor involving only a (typically much smaller) subset of the variables. To fully specify a PGM, in addition to the graphical structure, it is necessary to know the value that each factor takes for each combination of variable assignments. This is typically specified via a set of parameters. Thus, defining a PGM involves choosing a \emph{structure} (typically inspired by the underlying phenomenon being modeled) and estimating its \emph{parameters} (typically chosen to produce the best possible fit to a dataset of interest, according to some fitness criterion, such as maximum likelihood). In this work we assume that the graphical structure has already been specified and concern ourselves with the second part, parameter estimation, also known as learning or training of PGMs.

\subsection{Generative vs Discriminative PGMs}

Generative\footnote{Literature is not always consistent about the meaning of ``generative'' models, we follow \citet{ng2002discriminative}. In particular, when we call a model ``generative'' we do not imply that there is a simple procedure to sample from it.} PGMs model the joint PDF of data, whereas discriminative PGMs model the PDF of a subset of ``output'' (target) variables given the ``input'' (evidence) variables.

An advantage of generative PGMs is that we can train the model once on all the data and then, at test time, decide which variables should act as evidence and which variables should act as targets, obtaining valid answers without retraining the model. Furthermore, we can also decide at test time that some variables fall in neither of the previous two categories (unobserved variables), and the model will use the rules of probability to marginalize them out.

This is in contrast with discriminative models, such as multilayer perceptrons (MLPs), in which the set of input and output variables needs to be specified at training time and cannot be changed after the fact without retraining. If there is a need for more than a single evidence/target/unobserved configuration of the variables at test time (query), then a separate MLP needs to be trained for each configuration. In the extreme case in which any query is possible, the number of required MLPs would be exponential in the number of variables of the model, making this approach infeasible. Another disadvantage of the discriminative setup is that each query is trained for independently, despite being related tasks. However, in settings where the query is known a priori and fixed, discriminative models can work well.

\subsection{Directed vs Undirected PGMs}

In the fully observed case, the log-likelihood is a concave function of the natural parameters and its maximum likelihood (ML) estimate should in principle be easy to find. This is indeed the case for directed models, in which each factor can be fit to the data independently, and even has a closed form solution for fully parametric factors. In contrast, undirected PGMs are unnormalized and computing the normalization constant (the partition function) is intractable in general. This partition function depends on the parameters of the PGM, and thus needs to be taken into account during the learning process. The presence of hidden variables further complicates the ML learning problem, which is no longer concave and can have multiple local maxima.

The choice between directed and undirected models is usually motivated by the nature of the data. A paradigmatic example of the success of undirected PGMs is the application of Markov random fields (MRFs) to image modeling \cite{li2009markov}. The pixels in an image have no natural ordering, so undirected models make the most sense. When the MRF is used to describe a ``latent'' pixel space from which the actual pixels are generated through a noisy channel, hidden variables emerge naturally. Image modeling also provides an example requiring flexible querying: when inpainting a new image, the subset of pixels to be impainted is only known at test time and cannot be prespecified during training.

\subsection{Learning in Undirected PGMs}

In order to maximize the log-likelihood of a model, the gradients of the partition function (the expectations of the sufficient moments) need to be computed. These could be obtained via Markov chain Monte Carlo (MCMC), but this approach is often too slow for large or complex models.

Practical approaches to learn fully observed undirected models resort instead to fitness functions other than the likelihood, thus sidestepping the computation of the partition function. Relevant examples in the literature are contrastive free energy minimization \cite{welling2005learning}, piecewise training \cite{sutton2005piecewise}, pseudo-moment matching \cite{wainwright2003tree}, or pseudo-likelihood maximization \cite{besag1975statistical,hyvarinen2006consistency, sutton2007piecewise}.

These alternative, tractable fitness functions are sometimes guaranteed to still recover the ``correct'' (maximum likelihood) parameters of the model as the number of data tends to infinity under some assumptions, but those assumptions rarely hold in practice. For instance, pseudo-moment matching and piece-wise training will recover the correct parameters only if the PGM is a tree \cite{wainwright2003tree} whereas pseudo-likelihood maximization requires the model space to contain the true generative process of the observed data (almost never the case in practice).

 When hidden variables are present, the additional problem of marginalizing them out makes learning even harder \cite{welling2005learning,kuleshov2017neural,wiseman2019amortized,advil}.  This is the case that we consider in this work.

\subsection{Learning a Worse PGM for Better Marginals}

As we saw above, practical methods for learning undirected PGMs return parameters that deviate from the ML estimations. Furthermore, since inference is also intractable in general, in practice we query PGMs by resorting to approximate inference. The errors caused by inference are then compounded with the modeling errors. This thought has motivated researchers to embrace the idea of actually learning an admittedly ``wrong'' PGM, but design the learning scheme to compensate for the inaccuracy of approximate inference, such that the two cancel each other out \cite{stoyanov2011empirical}. This is e.g. the motivation behind pseudo-moment matching as described in \citet{wainwright2006estimating}. In this way, we learn a worse PGM (as compared with the ML estimation) in the hope of obtaining a better approximations for the marginals. Obviously, this ``worse PGM'' must be paired with a concrete inference scheme. If we were to run a sophisticated MCMC sampler to evaluate the test likelihood of the model, the results would likely be catastrophic.

Even though the idea of learning an appropriately wrong model is appealing, it is still missing information about how we are going to query the model. To illustrate this, consider the approach in \citet{wainwright2006estimating}. The author chooses the ``wrong'' parameters in such a way that when inference is applied \emph{in the absence of additional evidence}, the right marginals are retrieved. While that specific choice of query (no evidence) seems pretty neutral, it is unlikely to reflect how the model is used in practice. For instance, if all queries were to provide evidence for all the variables minus a subset that is connected forming a tree (different for each query), we know that using the correct parameters is what would result in exact marginals. Conversely, using the correct parameters would fail to provide the right marginals for their selected ``no evidence'' query when the model is loopy.

Thus, to learn the parameters of a PGM in a way that compensates for the inaccuracies of inference, one needs to take into account how the model is going to be queried at test time, i.e., use a \emph{query distribution}. The expectation is that even a conservative choice, such as a uniform density over all queries, will already generalize better to new queries than optimizing for a single one as in previous works.

\subsection{Our Approach}

Our approach can be described simply at a high level: choose an inference algorithm that can perform marginal inference in the presence of unobserved variables, such as loopy belief propagation (LBP), choose a query distribution, and choose the parameters that maximize the accuracy of the inference algorithm under the query distribution. Because LBP can be unrolled over time into a feed-forward network, parameter optimization can be performed easily using automatic differentiation and stochastic gradient descent. Tools like PyTorch \cite{paszke2019pytorch} and TensorFlow \cite{tensorflow2015-whitepaper} make gradient computation straightforward.

Prior work unrolling inference into a
neural network (NN) includes \citet{domke2011parameter,yoon2019inference,lin2015deeply}. However, none of these approaches consider the flexible querying that PGMs are expected to exhibit, and instead fix a single set of inputs and outputs, so that the resulting machine operates in effect as a NN. The cited approaches use PGMs as a motivation for the NN structure, but once the NN is created, the PGM capabilities are not retained.

In contrast, we propose to unroll LBP into a NN that takes as input both the data and a mask describing the query. At training time, that mask will be drawn from the query distribution, randomly attributing variables to the roles of evidence or target, whereas at test time it is given by the query.

Our goal is to provide a systematic framework that turns any PGM into a single NN that (a) handles undirected PGMs with hidden variables; (b) supports flexible \emph{marginal} querying  without retraining; and (c) provides a simple mechanism for training and inference.

 % also handles unobserved variables

\section{Query Training (QT)}
\label{sec:qt}

Our approach takes an (untrained) PGM and unrolls it into a single NN to answer arbitrary queries. It then trains the weights of the NN using different types of queries. The resulting NN can be used at test time for flexible querying, as if it was a PGM. We call this approach query training (QT).

\subsection{Queries that Need to Be Answered}

Given a knowledge representation of some type (e.g. PGM) for the set of variables $\bm{x} = \{x_1,
\ldots,x_N\}$, we want to be able to compute conditional marginal probabilistic queries of the form
\begin{align}
\label{eq:querytype}
&p(x_\text{target} | \{x_i\}_{i\in\text{evidence}})\\
\nonumber&\forall~ \text{target} \in \{1, \ldots, N\} , ~\forall~\text{evidence} \subset \{1, \ldots, N\}
\end{align}
where $x_\text{target}$ is a single variable, and ``evidence'' is the subset of the remaining variables for which evidence is available. Any variables that do not correspond to the input nor the output are marginalized out in the above query.

Queries which do not fit Eq.~\eqref{eq:querytype} directly (e.g., the joint distribution of two output variables) could be decomposed into a combination of conditional marginal queries by using the chain rule of probability. Though we do not pursue this decomposition here, we want to emphasize that a system that is able to answer queries like Eq.~\eqref{eq:querytype} contains enough information to resolve any probabilistic query, with the number of queries being linear in the number of output variables\footnote{If the answers to these queries is approximate (as it will be the case with QT), different factorizations of a joint density will result in different approximations.}. Because of this, QT is designed to answer marginal queries like Eq.~\eqref{eq:querytype} only. In order to have a single system answer arbitrary queries, we follow approximate inference in PGMs.

% A brute force solution would be to consider each of these queries its own regression problem and train separate NNs. However, the number of different queries of this type is exponential in the number of variables, so this would be infeasible. Ideally, we would like to train a single NN that can be reconfigured to address different queries. To this end, we will follow approximate inference in PGMs.

\subsection{Approximate Inference in PGMs}

Probabilistic queries in PGMs are in most cases intractable, so approximations such as loopy belief propagation (LBP, \citeauthor{murphy2013loopy}) or variational inference (VI, \citeauthor{blei2017variational}) are used. These approximations are invariant to scale, so the computation of the partition function is \textit{not needed}.

LBP and VI solve an optimization problem iteratively, which is slow. To speed up the process, amortized inference can be used. A prime example of this are VAEs \cite{kingma2013auto}: a
learned function (typically an NN) is combined with the reparameterization trick \cite{rezende2014stochastic, titsias2014doubly} to compute the posterior over the hidden variables given the visible ones.
Although a VAE performs inference faster than VI optimization, a single predefined query can be solved. In contrast, LBP and VI answer arbitrary queries (albeit with more compute). Note that standard VAEs can only handle directed PGMs, and a sophisticated variational apparatus with multiple NNs \cite{advil} is required for undirected PGMs.

LBP and VI are closely connected, but behave differently. For instance, for tree PGMs, parallel LBP converges to the exact solution in a number of steps equal to the diameter of the tree \cite{murphy2013loopy}, whereas VI will in general take much longer to converge. In general, LBP tends to produce higher quality marginals in less iterations, so we will choose it over VI in this work.

\subsection{An Intuitive Description of Query Training}

LBP gives a recipe to compute any marginal query of the form of Eq.~\eqref{eq:querytype}, while sharing the same parameters for all queries. We can then consider LBP, unrolled over a fixed set of iterations, as our inference NN. This NN takes an additional input specifying the desired  query. Instead of starting with a trained PGM and generating the inference NN (which is certainly possible), we could generate a ``blank'' inference NN from an untrained PGM and then learn its (single set of) parameters by minimizing the cross-entropy (CE) between data and its predictions, averaged over the query distribution. We expect this to generalize to new data points and new queries never seen at training time. The intuition behind the existence of a single NN parameterization that approximately satisfies all the queries comes from the good results of running LBP on a correct PGM, which uses a single set of parameters. Note that the the same set of parameters (weights) are shared across all inference steps (layers).

Minimizing the CE between the training data and the query predictions %(even if those predictions were exact for the current model parameters)
with respect to the model parameters is \emph{not} equivalent to maximum likelihood learning.
%However, we have derived consistency results (see Supplementary Material) for general exponential family models, analogous to those of pseudolikelihood \cite{hyvarinen2006consistency}, showing that our CE loss is reasonable.

The computation of the partition function is sidestepped, so undirected PGMs can be used just as easily:  normalization is only required for the output variable of each query, i.e., in one dimension.

LBP is in general only approximate, so one cannot expect predictions to be exact or necessarily consistent (e.g., the query product $p(x=0|y=1)p(y=1)$ is not guaranteed to be identical to the query product $p(y=1|x=0)p(x=0)$, although the cost function will tend to make both similar). On the plus side, the training is free to select the parameters that produce the most precise inference over the training queries, as opposed to the best fit to the data in the ML sense, so this ``worse'' PGM can compensate for shortcomings in the LBP approximation resulting in better marginals.
\begin{figure*}[!htbp]
\centering
  \includegraphics[width=0.8\linewidth]{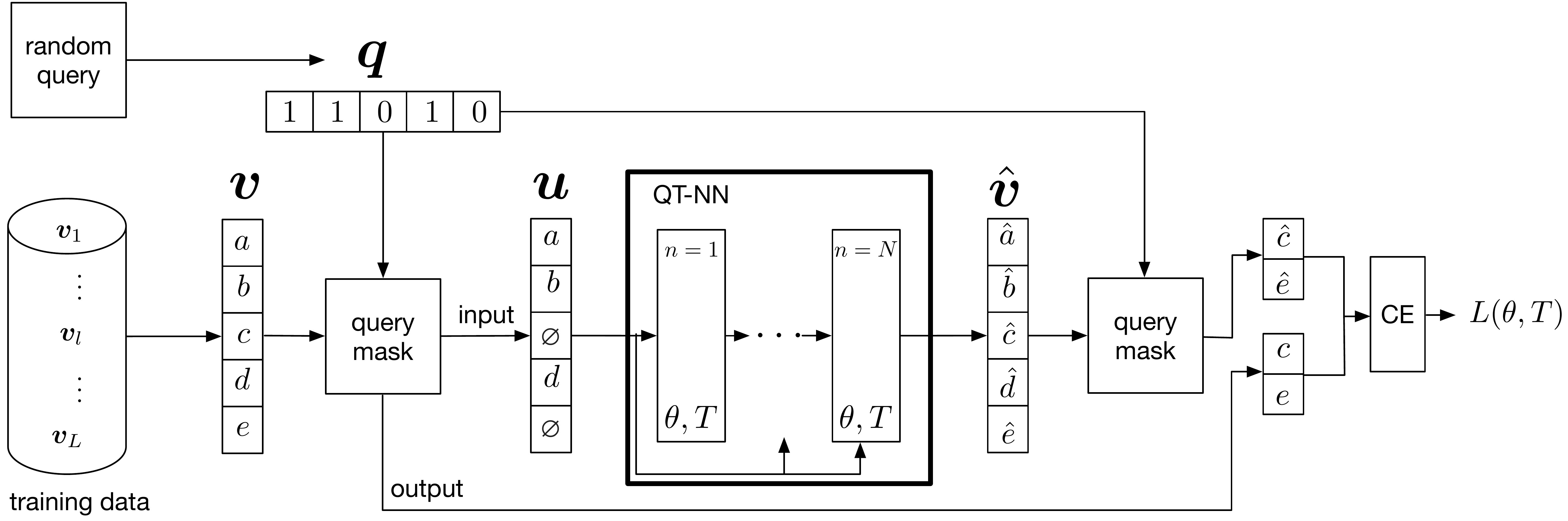}
  \caption{One step of query training. A random sample from the training data is split according to a random query mask in input and output dimensions. The input is processed inside the QT by $N$ identical stages, producing an estimation of the marginal probabilities from the sample. The loss function considers the cross-entropy between the true and estimated marginals. Training uses stochastic samples from both training data and training queries.}
  \label{fig:qt}
\end{figure*}

\subsection{Training the Inference Network}

The starting point is an unnormalized PGM parameterized by $\theta$. Its probability density can be expressed as  $p({\bm x}; \theta) = p({\bm v}, {\bm h}; \theta) \propto \exp(\phi({\bm v}, {\bm h}; \theta))$ , where ${\bm v}$ are the visible variables available in our data and $\bm{h}$ are the hidden variables. A query is a binary vector $\bm{q}$ of the same dimension as $\bm{v}$ that partitions the visible variables in two subsets: one for which (possibly soft) evidence is available (inputs) and another whose marginal probability we want to estimate (outputs).
Note that we want to compute the marginal of each selected output given all the inputs, independently, %I.e., we want to compute as many independent queries as output variables are specified by $\bm{q}$.
that is, for $\mathcal{S} = \{i: q_i = 1\}$, we want to compute all the marginal queries $p(x_i | \{ x_j \}_{j \in \mathcal{S}}), ~ \forall i \notin \mathcal{S}$.

The query-trained neural network (QT-NN) follows from specifying a graphical model $\phi({\bm v}, {\bm h}; \theta)$, a temperature $T$ and a number of inference timesteps $N$ over which to run parallel BP. The general equations of the QT-NN are given in the next Subsection.

In Fig.~\ref{fig:qt}, a QT-NN takes as input a sample $\bm {v}$ from the dataset and a random query mask $\bm {q}$. The query $\bm{q}$ blocks the network from accessing the ``output'' variables, and instead only offers access to the ``input'' variables. The variables assigned to inputs and and outputs change with each query $\bm{q}$ drawn. The QT-NN produces as output an estimate of the marginal probabilities $\hat{\bm {v}}$ for the whole input sample. Obviously, we only care about how well the network estimates the variables that it did not see at the input. So we measure how well $\hat{\bm {v}}$ matches the correct $\bm {v}$ in terms of cross-entropy (CE), but only for the variables that $\bm{q}$ regards as ``output''.

Taking expectation wrt $\bm {v}$  and $\bm {q}$, we get $L(\theta, T) = \mathbb{E}_{\bm{v},\bm{q}}[\operatorname{CE}_{\bm{q}}(\bm{v}, \hat{\bm{v}})]$, the loss function that we use to train the QT-NN, where the estimated visible units are given by $\hat{\bm{v}} = \operatorname{QT-NN}(\bm{v}, \bm{q}; \theta, T)$.

We minimize this loss wrt $\theta, T$ via SGD, sampling from the training data and some query distribution. The query distribution should follow our expected use of the system at test time, and can be set to uniform by default. The number of QT-NN layers $N$ is fixed a priori. %Because we are not only training over data samples, but also over possible queries,
%we are training over possible queries,
%We term this approach query training (QT).

One can think of the QT-NN as a more flexible version of the encoder in a VAE: instead of hardcoding inference for a single query (normally, hidden variables given visible variables), the QT-NN also takes as input a mask $\bm {q}$ specifying which variables are observed, and provides inference results for unobserved ones. Note that $\bm{h}$ is never observed, and instead approximately marginalized over by BP.

\subsection{Unrolling BP into a QT-NN}
\label{sec:bptonn}

For a given set of graphical model parameters $\theta$ and temperature $T$ we derive a feed-forward function that approximately resolves arbitrary inference queries by unrolling the parallel BP equations for $N$ iterations. First, we combine the available evidence $\bm{v}$ and the query $\bm{q}$ into a set of unary factors. Unary factors specify a probability density function over a variable. Therefore, for each dimension inside $\bm{v}$ that $\bm{q}$ labels as ``input'', we provide a (Dirac or Kronecker) delta centered at the value of that dimension. For the ``output'' dimensions and hidden variables $\bm{h}$ we set the unary factor to an uninformative, uniform density. Finally, soft evidence, if present, can be incorporated through the appropriate density function. The result of this process is a unary vector of factors $\bm{u}$ that only contains informative densities about the inputs and whose dimensionality is the sum of the dimensionalities of $\bm{v}$ and $\bm{h}$. Each dimension of $\bm{u}$ will be a real number for binary variables (which we encode in the logit space), and a full distribution in the general case.

Once $\bm{v}$ and the query $\bm{q}$ are encoded in $\bm{u}$, we can unroll parallel LBP over iterations as an NN with $N$ layers, i.e., the QT-NN. To simplify notation, let us consider a PGM that contains only pairwise factors. By mapping the messages to the log-space, the predictions of the QT-NN and the messages from each layer to the next can be written as
\begin{align*}
m_{ij}^{(0)} &= 0 ~~~~~ m_{ij}^{(n)} = f_{\theta_{ij}}\Big(\theta_i + u_i + \sum_{k\neq j} m_{ki}^{{(n-1)}} ; T\Big)\\
\hat{\bm{v}}_i &= \operatorname{softmax}\Big(\theta_i + u_i + \sum_k m_{ki}^{(N)}\Big)\\
\bm{m}^{(0)} &= 0~~~~~
\bm{m}^{(n)} = f_\theta(\bm{m}^{(n-1)}, \bm{u}; T)~~~~~
\hat{\bm{v}} = g_\theta(\bm{m}^{(N)}, \bm{u}),
\end{align*}
where the last row express the same equations as the other two, but in vectorized format.

Here $\bm{m}^{(n)}$ collects all the messages\footnote{For a fully connected graph, the number of messages is quadratic in the number of variables, showing the advantage of a sparse connectivity pattern, which can be easily encoded in the PGM architecture.} that exit layer $n-1$ and enter layer $n$. Messages have direction, so $m_{ij}^{(n)}$ is different from $m_{ji}^{(n)}$\footnote{In the general case, $m_{ij}^{(n)}$ and $\theta_i$ are vectors. We only encode matrix in bold, to represent aggregate messages}. Observe how the input term $\bm{u}$ is re-fed at every layer. The output of the network is a belief $\hat{\bm{v}}_i$ for each variable $i$, which is obtained by a softmax in the last layer.  These equations simply correspond to unrolling LBP over iterations, with its messages encoded in log-space.

The portion of the parameters $\theta$ relevant to the factor between variables $i$ and $j$ is represented by $\theta_{ij} = \theta_{ji}$, and the portion that only affects variable $i$ is contained in $\theta_i$. Observe that all layers share the same parameters. The functions $f_{\theta_{ij}}(\cdot)$ are directly derived from $\phi(\bm{x};\theta)$ using the BP equations, and therefore inherit its parameters. Finally, parameter $T$ is the ``temperature'' of the message passing, and can be set to $T=1$ to retrieve the standard sum-product belief propagation or to 0 to recover max-product belief revision. Values in-between interpolate between sum-product and max-product and increase the flexibility of the NN.
Refer to the Supplementary Material for the precise equations obtained when applied to the three popular undirected models used in our experiments.

\subsection{Connection with Pseudo-likelihood}
If the  query distribution is uniform over all queries with exactly one target variable (the rest being evidence), and if there are no hidden variables, the above cost function reduces to the pseudo-likelihood (PL) of \cite{besag1975statistical}. Other query distributions can be seen as composite likelihoods.
However, unlike PL or composite likelihoods, QT can approximately marginalize hidden variables and does not aim to learn the true parameters of the model, but instead optimizes them for best performance at inference time.
\begin{table*}[!htbp]
\centering
\smallskip
\resizebox{0.8 \textwidth}{!}{%
  \begin{tabular}{lcccccccccc}
  \toprule
  Method & Model & \bfseries Adult & \bfseries Conn4 & \bfseries Digits & \bfseries DNA& \bfseries Mushr& \bfseries NIPS& \bfseries OCR& \bfseries RCV1& \bfseries Web \\
  \midrule
AdVIL-BP  & RBM &  0.224  & 0.248  & 0.530  & 0.778  & 0.192  & 0.795  & 0.470  & 0.475  & 0.142 \\
AdVIL-Gibbs  &RBM &   0.229  & 0.238  & 0.493  & 0.782  & 0.218  & 0.797  & 0.471  & 0.477  & 0.163 \\
PCD-BP  & RBM &  0.215  & 0.285  & 0.530  & 0.763  & 0.159  & 0.801  & 0.428  & 0.457  & 0.140 \\
PCD-Gibbs  & RBM &  0.218  & 0.288  & 0.516  & 0.765  & 0.159  & 0.804  & 0.427  & 0.458  & 0.144 \\
QT (Ours)  & RBM &  \textbf{0.167}  & \textbf{0.148}  & \textbf{0.472}  & 0.766  & \textbf{0.124}  & \textbf{0.787}  & \textbf{0.377}  & \textbf{0.452}  & \textbf{0.133} \\
  \midrule
AdVIL-BP  & DBM &  0.242  & 0.364  & 0.578  & 0.805  & 0.224  & 0.816  & 0.490  & 0.480  & 0.146 \\
AdVIL-Gibbs  &DBM &   0.244  & 0.299  & 0.529  & 0.560  & 0.690  & 0.934  & 0.923  & 0.568  & 0.195 \\
QT (Ours)  &DBM &   \textbf{0.169}  & \textbf{0.146}  & \textbf{0.471}  & 0.765  & \textbf{0.124}  & \textbf{0.787}  & \textbf{0.379}  & \textbf{0.453}  & \textbf{0.133} \\
\bottomrule
  \end{tabular}}
\caption{Comparison of QT-NN, PCD and AdVIL on RBM and DBM. Measurements are NCE, lower is better. QT achieves significantly lower NCE for most datasets compared to the state-of-the-art. Code: \url{https://github.com/vicariousinc/query_training}.}
\label{tab:rbmdbm_results}
\end{table*}

\section{Experiments}
\label{sec:experiments}

Early works in learning undirected PGMs relied on contrastive energies \cite{hinton2002training,welling2005learning}. More recent approaches are NVIL \cite{kuleshov2017neural} and AdVIL \cite{advil}, with the latter being regarded as superior. We will use it as our main benchmark.

We will test QT as an approach for learning and querying in undirected PGMs, comparing it with AdVIL on 3 different types of undirected PGMs (using both discrete and continuous variables) and a total of 10 datasets. These models and datasets are exactly the ones used in AdVIL's paper \cite{advil}. We also apply QT to a challenging grid MRF.

\subsection{Comparison with AdVIL}
All  models are trained using uniform random query distribution, i.e., in each SGD step, each variable is independently assigned as input or output with 0.5 probability. We report the test normalized cross-entropy (NCE) in bits to measure generalization to new data \emph{and new queries}. We normalize the number of predicted variables, so the NCE is the average CE per-variable. The query distribution at test time is the same as the training one, but the actual queries are with very high probability unseen in training. Experiments are run on a single Tesla V100 GPU.

The QT-NN equations for the models in this section (simply unrolling LBP on each PGM) and code to reproduce our results can be found in the Supplementary Material.

\subsubsection{Restricted Boltzmann Machine (RBM)}\label{sec:rbm}

We first test QT on which is arguably the simplest undirected model with hidden variables, the RBM.
The log-probability of an RBM\footnote{This parameterization is in one-to-one correspondence with the standard and results in simpler QT-NN equations as shown in the Supplementary Material. Most readers can ignore this detail.} is proportional to $\phi({\bm v}, {\bm h}; \theta) = 2{\bm h}^\top W{\bm v} + {\bm h}^\top (\bm{c}_H-W\bm{1}_V) + {\bm v}^\top (\bm{c}_V-W^\top1_H)$.
We use exactly the same datasets and preprocessing used in the AdVIL paper, with the same hidden layer sizes, check \citet{advil} for further details. Since all the variables are binary, a trivial uniform undirected model would result in an NCE of 1 bit.
%This can be used as a reference for interpretation.

We also include results from persistent contrastive divergence (PCD), which is known to be very competitive for RBM training \cite{tieleman2008training,marlin2010inductive}. In fact, AdVIL does not do much better than PCD on this model.

Computing the test NCE for QT only requires to running the trained QT-NN on test data and evaluating its loss function. PCD and AdVIL, however, do not provide any mechanism to solve arbitrary inference queries, so we needed to resort to Gibbs sampling in the learned model, which is much slower. Alternatively, we also tried copying the RBM weights learned by PCD and AdVIL into the QT with $T=1$ and report those results as PCD-BP and AdVIL-BP.

For AdVIL we use the code provided by the authors. For PCD and QT the validation set is used to choose the learning rate and for early stopping, separately for each dataset. The learning rates considered are $\{0.03, 0.1, 0.3, 1, 3\}$ for PCD and $\{0.001, 0.003, 0.01, 0.03\}$ for QT. For PCD we use {\tt scikit-learn} \cite{scikit-learn}. For QT we unfold BP in $N=10$ layers and use ADAM \cite{paszke2019pytorch, kingma2014adam} with minibatches of size 500. The $T$ parameter is learned during training. The results are shown in Table \ref{tab:rbmdbm_results}. Results average 5 independent runs.

QT produces significantly better results for most datasets (marked in boldface), showing that it has learned to generalize to new probabilistic queries on unseen data.

\subsubsection{Deep Boltzmann Machine (DBM)}

For our second experiment, we use a slightly more sophisticated undirected PGM, a DBM with two hidden layers.
The log-probability of a DBM\footnote{Previous footnote also applies here.} is proportional to
$\phi({\bm v}, {\bm h}; \theta) =
2{\bm h_2}^\top W_{H_2 H_1}{\bm h_1} + 2{\bm h_1}^\top W_{H_1 V}{\bm v} +
{\bm h_2}^\top (\bm{c}_{H_2}-W_{H_2 H_1}\bm{1}_{H_1})  +
{\bm v}^\top (\bm{c}_V-W_{H_1 V}^\top1_{H_1}) +
{\bm h_1}^\top (\bm{c}_{H_1}-W_{H_2 H_1}^\top\bm{1}_{H_2} - W_{H_1 V}\bm{1}_V)$.
 The datasets are the same as for the RBM. The DBM structure (number of units in each hidden layer) for each dataset is identical to that of \citeauthor{advil}, and we reuse exactly the same process and parameters as in the previous experiment. The results are summarized in Table \ref{tab:rbmdbm_results}.

QT produces the best performance (marked in boldface) on 8 out of the 9 tested datasets. It is interesting to note that, although inference becomes more challenging for DBM than for RBM (as demonstrated by the generally higher NCEs on most datasets for AdVIL-BP and AdVIL-Gibbs), the performance of QT remains essentially unchanged for DBM as compared with RBM. This suggests that QT not only learns to generalize to new  probabilistic queries on unseen data, but also remains highly effective for models where inference becomes more challenging.

\subsubsection{Gaussian Restricted Boltzmann Machine (GRBM)}
\label{sec:grbm}
Finally, we compare QT with AdVIL on learning a GRBM on the Frey faces dataset.
The log-probability of a GRBM is proportional to $\phi({\bm v}, {\bm h}; \theta) = -\frac{1}{2 \sigma^2} \|{\bm v}- {\bm b}\|^2 + {\bm c}^\top {\bm h} + \frac{1}{\sigma} {\bm h}^\top W {\bm v}$. The hidden units are binary and the visible units are continuous. We follow \citeauthor{advil}, and fix $\sigma=1$. The dimensions of ${\bm h}$ and ${\bm v}$ are respectively 200 and 560 (corresponding to a $28\times 20$ image).

We train AdVIL using the provided authors' code. For QT we unfold BP in $N = 50$ layers and use ADAM with learning rate $5\times10^{-3}$. In the GRBM, BP will send continuous messages to the visible units, and we follow the standard practice of expectation propagation \cite{minka2001expectation} of characterizing those messages as Gaussians. This results in deterministic and differentiable message updates. For further details, see the Supplementary Material.

Since no quantitative results are provided in \cite{advil}, we create our own by using the preexisting train-test split in the dataset of 1572 and 393 images, respectively. Since there is no validation split, for QT we simply stop training after 50 epochs (training performance has plateaued and we see no signs of overfitting). For AdVIL, we train for 100,000 iterations and save a checkpoint every 100 iterations. We observe overfitting, so we decided to give the competing method an additional advantage and we report the results of the checkpoint that provides the best test set performance.  Like in the previous experiments, the results for AdVIL are obtained both by using Gibbs sampling and by transferring its learned weights to a QT in which we run $N = 50$ BP iterations.
The NCE for each of the models are Advil-Gibbs: $1.545$, Advil-BP: $1.542$, and QT: $\mathbf{1.503}$. The NCE of a trivial independent Gaussian model using the empirical mean and variance of the training pixels is $1.909$.
Again, QT produces better results than AdVIL, showing that it can be applied to continuous PGMs.

{\bf Testing on significantly different queries.} Ideally, the QT-NN should be trained under a query distribution that matches the query statistics at test time. However, even using a conservative, uniform distribution should already be better than optimizing for a single fixed query. In order to assess this we use the GRBM trained on uniform queries and test it on significantly different ones, asking to complete contiguous patches of $5 \times 5$ pixels given the rest of the image. The NCE for each of the models are: Advil-Gibbs: $1.525$, Advil-BP: $1.530$, and QT: $\mathbf{1.493}$. The NCE of the trivial model is $1.889$. Qualitatively, we find that QT is able to produce image completions that are almost indistinguishable from the original images (shown in Fig \ref{fig:nips_figures}a).

This shows the PGM-like flexibility of QT, answering queries on blocks that it was never trained for. Of course, prediction will degrade as the statistics of the test queries depart further from the training ones.

\subsection{Grid Markov Random Field (GMRF)}
We consider the challenging problem of using QT to learn an 8-connected, grid-arranged Markov random field (MRF) \emph{with hidden variables}, as shown in Fig.~\ref{fig:nips_figures}b. Although the models explored so far also had hidden variables, they did not have direct connections---as it is the case now---which make the model more loopy and learning more challenging. Grid MRFs are often used in image processing applications \cite{li2009markov}, but the MRF variables are always observed when learning the factor parameters. In our case, the MRF variables are hidden, and they emit the pixel labels through a noisy channel, with multiple hidden states mapping to the same pixel labels.

To the best of our knowledge, QT is the first method that can efficiently learn the full parameterization of an 8-connected grid MRF without direct access to the MRF variables (which are hidden and only visible through a noisy channel). Although irrelevant for our purpose of learning a challenging undirected PGM, the proposed model is a simple incarnation of visual neuroscience principles for foreground--background segmentation. Further details about the  model and its neuroscience motivation are presented in the Supplementary Material.

\subsubsection{The Border Ownership Dataset, Model, and Task}

\begin{figure*}[!htbp]
    \centering
    \includegraphics[width=0.8\textwidth]{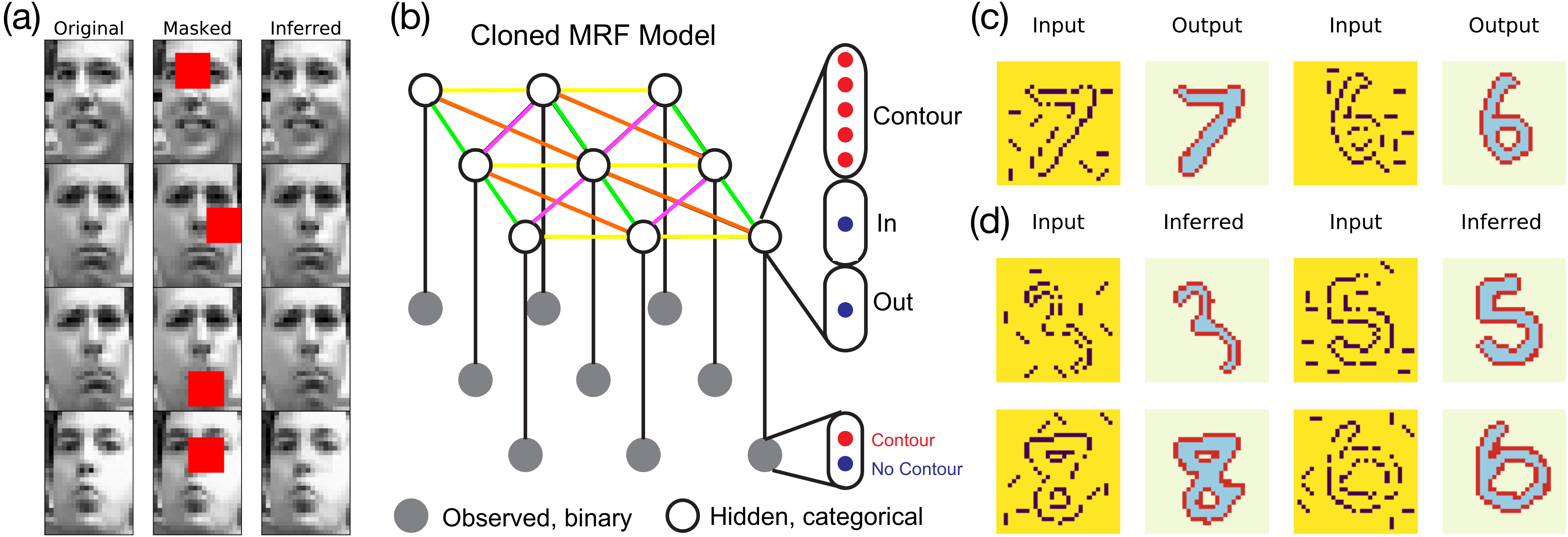}
    \caption{(a) QT can accurately complete masked regions of an image even though it was trained for significantly different queries. (b) An 8-connected cloned Markov random field. Identical factors are shown using the same color. Actual size is $30\times 30$. (c) Two examples of training pairs: a noisy input digit and its corresponding ground truth segmentation (d) Test data: noisy input digits and their inferred segmentation by the QT-NN (which is obtained unrolling the GMRF model).}
    \label{fig:nips_figures}
\end{figure*}

The border ownership dataset is provided with the Supplementary Material and is derived from the MNIST dataset\footnote{The MNIST dataset can be found at \url{http://yann.lecun.com/exdb/mnist/}.} \cite{lecun2010mnist}. It is structured as pairs of noisy contour images and \texttt{CONTOUR}-\texttt{IN}-\texttt{OUT} labels. Two examples are displayed in Fig.~\ref{fig:nips_figures}c. The contours are missing with probability 0.2, whereas each image incorporates 8 spurious random edges of length 3 pixels. Each image is of size $30 \times 30$ pixels. The images have one-to-one correspondence with the MNIST dataset, so 60,000 images are available for training and 10,000 images are used for testing.

The structure of the MRF is shown in Fig.~\ref{fig:nips_figures}b. Each node is a categorical variable with 66 hidden states, 64 correspond to the label \texttt{CONTOUR}, 1 to \texttt{IN}, and 1 to \texttt{OUT}. The vertical connections are a noisy channel. There are only 4 distinct pairwise factors (different colors in Fig.~\ref{fig:nips_figures}b). The task is to recover the hidden labels from the noisy image. Observe that the incomplete contours and the spurious edges make the task of foreground--background segmentation non-trivial. Second, observe that the labels do \emph{not} provide the hidden states. In particular, which of the 64 clones of \texttt{CONTOUR} is appropriate for each pixel is unknown, and the use of multiple clones is required to properly solve the task, since the potentials are only local pairwise connections and long-range information is needed.

\subsubsection{Results}
\begin{table}[b!]
    \centering
    \resizebox{.3\textwidth}{!}{
        \medskip
        \begin{tabular}{lcc}
        \toprule
        Method & \textsc{IOU} & \textsc{NCE}\\
        \midrule
        \textsc{QT (Ours)} & $\mathbf{96.97}$ & $\mathbf{0.0398}$ \\
        \textsc{Random} & $16.39$ & $1.00$ \\
        \bottomrule
        \end{tabular}
    }
    \caption{\small{Results for QT and a random baseline for GMRF on the test border ownership dataset.}}\smallskip
    \label{tab:CMRF}
\end{table}
The model is trained using QT. The input and output variables in this case are not randomized, but fixed throughout training and testing: the evidence is always the noisy binary image and the target is always the noiseless ternary segmentation. We unroll BP for $N=15$ layers, use ADAM with a learning rate of $10^{-2}$ with minibatches of 50 images and run learning for 10 epochs on a single Tesla V100 GPU. The temperature parameter is fixed $T=1$ throughout learning. %We chose these parameters simply by looking at the training loss; the amount of training data was so large compared with the number of weights that no overfitting was likely.
The results of segmentation from noisy test data are shown on Fig.~\ref{fig:nips_figures}d for several example digits. Pixels decoded as \texttt{CONTOUR}, \texttt{IN}, \texttt{OUT} are respectively in red, pale blue, and pale yellow. Qualitatively, the recovery looks almost perfect. Quantitatively, Table \ref{tab:CMRF} presents the intersection over union (IoU) between the estimated and real foreground\footnote{For the purpose of this metric, we consider foreground those pixels labeled (or estimated) as either \texttt{IN}, \texttt{CONTOUR}. Higher is better.}, and the NCE of a GMRF trained with QT and with a random baseline. QT achieves a nearly perfect digit recovery.
%Quantitatively, we measured the intersection over union (IoU) between the estimated and real foreground\footnote{For the purpose of this metric, we consider foreground those pixels labeled (or estimated) as either \texttt{IN}, \texttt{CONTOUR}. Higher is better.}, and the NCE to be 96.97\% and 0.0398, respectively. For contrast, a random model that assigns pixels to \texttt{IN} or \texttt{OUT} with 0.5 probability would yield 16.39\% and 1.00, respectively.

We have tried to train the GMRF using other alternatives without luck. In particular, AdVIL requires designing three new encoder networks and one decoder network for this model. Our network designs have failed to produce any meaningful results in a reasonable amount of time.

\section{Discussion and Future Work}
\label{sec:future}

Query training is a general approach to parameter learning in PGMs when these need to support flexible querying. It offers the following advantages:
% \begin{enumerate}
(a) simple to implement (just differentiable LBP);
(b) applicable to any PGM, even undirected ones with hidden variables;
(c) it sidesteps the estimation of the partition function;
(d) arbitrary inference is built in (e.g., AdVIL needed separate Gibbs sampling to answer our queries);
(e) it adapts learning to inference according to a \emph{query distribution}, ignored in previous works, which can be set to uniform and already provide reasonable generalization capabilities.
% \end{enumerate}

Why does QT generalize to new queries? The worry could be that only a small fraction of the exponential number of potential queries is seen during training. The \emph{existence} of a single inference network that works well for many different queries follows from the existence of a single, correct PGM in which LBP can approximate inference reasonably well for arbitrary queries. The \emph{discoverability} of such a network from limited training data is not guaranteed. However, there is hope for it, since the amount of training data required to adjust the model parameters should scale with the number of these, and not with the number of potential queries. Just like training data should come from the same distribution as test data, the training queries must come from the same distribution as the test queries to avoid ``query overfitting''.

Interesting avenues for exploration are model sampling and other inference mechanisms, such as VI.

\bibliography{query_training}{}

\end{document}

\begin{appendices}

\section{Hidden activations learned by the 8-connected GMRF}

\begin{figure}[h!]
    \centering
    \includegraphics[width=0.84\textwidth]{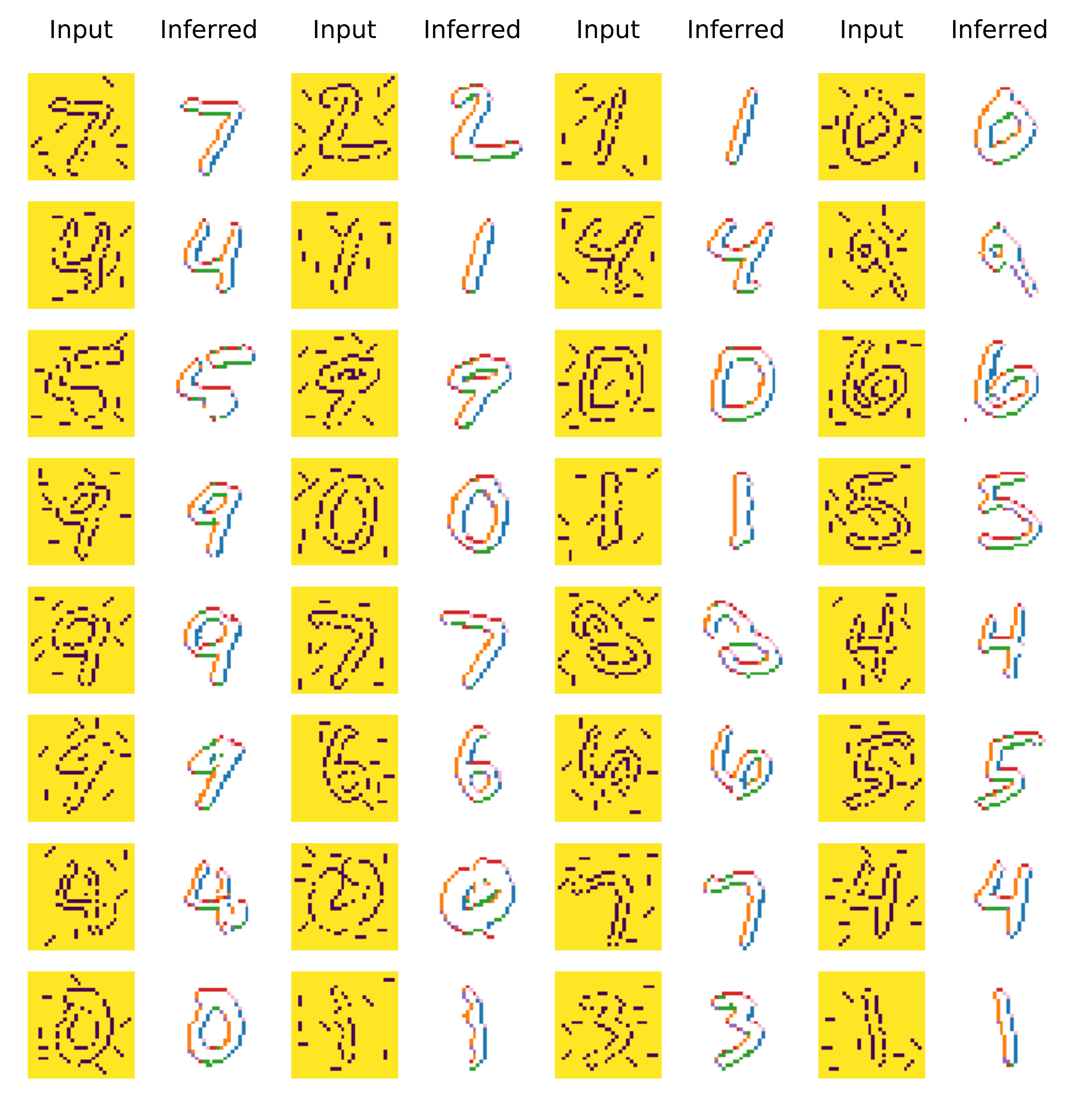}
    \caption{Best viewed on screen with zoom. Each color corresponds to a different inferred hidden contour clone. The model has learned \emph{with no supervision} to capture the local orientation of the pixel (which also reveals on which side the foreground is) as the best way to solve the denoising task.}
    \label{fig:bomnist_results}
\end{figure}

\section{Visual neuroscience and model details for the GMRF}

\begin{figure}[h!]
    \centering
    \includegraphics[width=.5\textwidth]{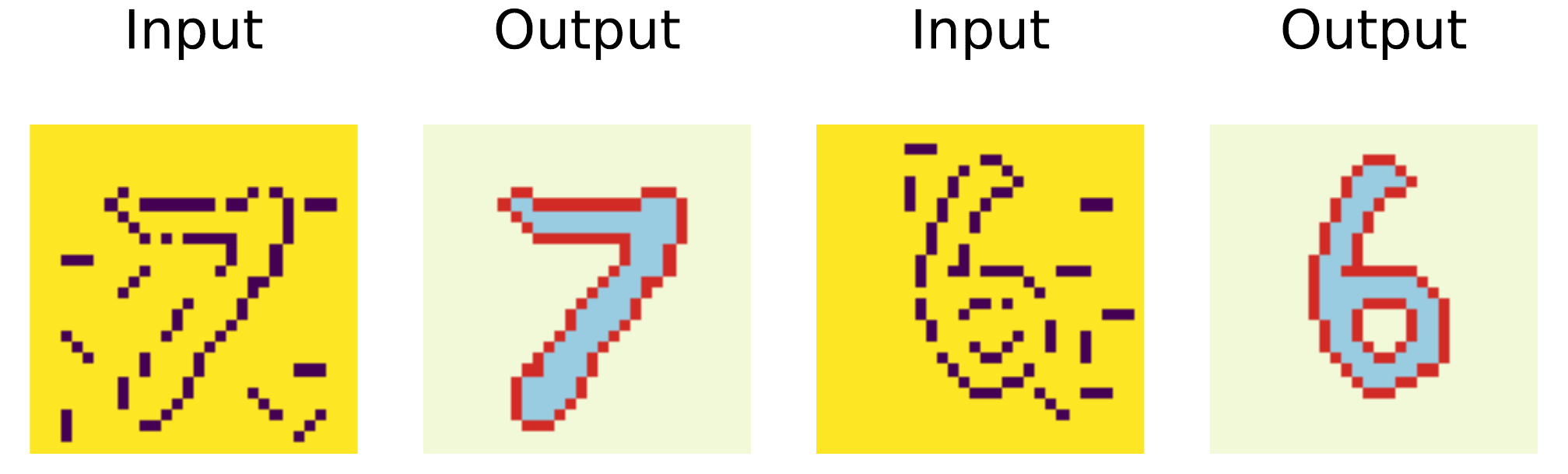}
    \caption{Two examples of training pairs: a noisy input digit and its corresponding ground truth segmentation}
    \label{fig:bomnist}
\end{figure}

\begin{figure}[h!]
    \centering
    \includegraphics[height=2in]{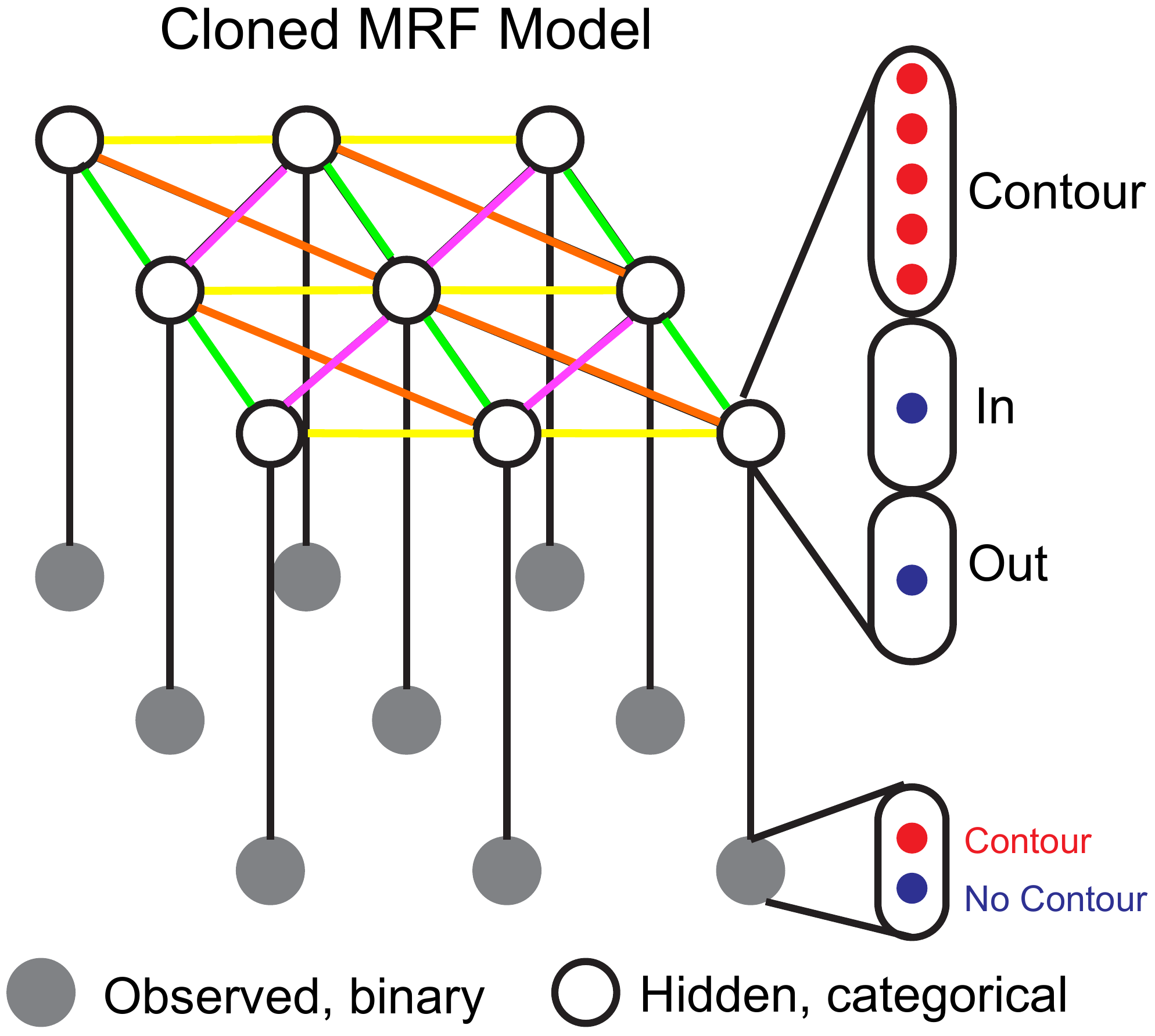}
    \caption{An 8-connected grid Markov random field. Identical factors are shown using the same color. Actual size, $30\times 30$.}
    \label{fig:gmrf}
\end{figure}

The purpose of this model is to perform some rudimentary foreground--background segmentation from noisy cues of the edges of the foreground, see Fig.~\ref{fig:bomnist}. There is abundant literature supporting border ownership as mechanism for foreground--background segmentation  \citep{o2009short, o2013remapping, zhaoping2005border, zhou2000coding},
and the use of ``clones'' for representing higher-order dependencies \citep{hawkins2009sequence,xu2016representing,cormack1987data}.
A simplified model based on these principles is presented in Fig.~\ref{fig:gmrf}. It has two types of variables, pixel labels (white, hidden, categorical with 66 categories) and pixel intensities (gray, observed, binary). The pixel labels categorize a pixel as belonging to the \texttt{OUT} (one category), \texttt{IN} (one category), or \texttt{CONTOUR} (64 distinct categories) of the foreground surface. In the noiseless case, the emission factors (vertical) are deterministic: the \texttt{IN} and \texttt{OUT} pixel labels produce pixel intensity 0, whereas the 64 remaining categories (\texttt{CONTOUR}) produce pixel intensities 1. The point of having 64 apparently identical categories, all of which generate a contour (these are the ``clones''), is that those hidden labels can specialize and learn higher order properties of the contour, such as its orientation and the side of the foreground on which they are located. Thus, a pixel that is part of a horizontal contour will have a different pixel label than another pixel that is part of a vertical contour. Now, a \texttt{CONTOUR} pixel whose hidden label is ``bottom horizontal'' is likely to turn on other  ``bottom horizontal''  pixels on its left and right and \texttt{IN} pixels on top of it. This allows to represent the long-range interactions required for foreground--background segmentation using only pairwise factors.
Note that these labels are just a possible specialization that training can discover, these are not available! The system contains unsupervised hidden variables. Due to its arrangement on a grid, we call this model the grid MRF (GMRF).

 \paragraph{Model details: }

First, let us consider the lateral connections between pixel labels in Fig.~\ref{fig:gmrf}. The model is fully convolutional, meaning that there are only 4 different types of potentials: up-down (green), left-right (yellow), principal diagonal (orange) and secondary diagonal (purple). These are color-coded in Fig.~\ref{fig:gmrf}, and extend in every direction to accommodate any GMRF size. This means that the model will perform the same segmentation regardless of where in the image the surface is presented (barring edge effects). These factors connect pairs of categorical variables with 66 values, are fully parametric, and are learned by QT.

The vertical connections (emission factors) are either known a priori or easy to estimate, so they are fixed during QT learning. In the noiseless case, they deterministically map to 0 or 1 as described above. In the noisy case, we have 3 noise parameters that determine the probability of generating a 1 conditioned on the label type being \texttt{CONTOUR}, \texttt{IN} or \texttt{OUT}. If we have access to ground truth segmentations and noisy images (as it is the case in our training data), those 3 parameters can be estimated in closed form.

To handle segmentation from noisy images, it is useful to think of this model as having both the noisy emission factors (producing the noisy image) and noiseless emissions factors that emit the segmentation categories: \texttt{CONTOUR}, \texttt{IN} and \texttt{OUT}. We will condition on the first and target the second.

\section{Details of experimental results}

\subsection{Restricted Boltzmann machine (RBM)}
\begin{table*}[!htbp]
\centering
\caption{Standard deviation of the mean for RBM results in the main paper}
\label{tab:rbm_results}
\smallskip
\resizebox{\textwidth}{!}{%
  \begin{tabular}{lcccccccccc}
  \toprule
  Method & \bfseries Adult & \bfseries Conn4 & \bfseries Digits & \bfseries DNA& \bfseries Mushr& \bfseries NIPS& \bfseries OCR& \bfseries RCV1& \bfseries Web \\
  \midrule
AdVIL-BP  & \num{4.55e-4}  & \num{7.44e-5}  & \num{1.66e-3}  & \num{2.22e-5}  & \num{4.36e-4}  & \num{7.9e-4}  & \num{2.83e-3}  & \num{1.48e-4}  & \num{2.29e-4} \\
AdVIL-Gibbs  & \num{7.84e-4}  & \num{4.98e-3}  & \num{1.29e-3}  & \num{8.71e-4}  & \num{4.06e-3}  & \num{6.77e-4}  & \num{8.55e-4}  & \num{1.13e-4}  & \num{7.32e-3} \\
PCD-BP  & \num{1.52e-3}  & \num{2.88e-3}  & \num{1.18e-3}  & \num{6.86e-4}  & \num{2.74e-3}  & \num{2.48e-3}  & \num{9.78e-4}  & \num{1.85e-4}  & \num{1.56e-4} \\
PCD-Gibbs  & \num{1.86e-3}  & \num{2.58e-3}  & \num{8.09e-4}  & \num{9.68e-4}  & \num{3.2e-3}  & \num{2.07e-3}  & \num{1.28e-3}  & \num{1.94e-4}  & \num{2.24e-4} \\
QT (Ours)  & \num{3.26e-4}  & \num{7.34e-4}  & \num{5.77e-4} & \num{8.45e-4}  & \num{1.68e-3}  & \num{6.37e-5}  & \num{1.09e-3}  & \num{3.06e-4}  & \num{2.06e-4} \\
\bottomrule
  \end{tabular}}
\end{table*}

\subsection{Deep Boltzmann machine (DBM)}

\begin{table*}[!htbp]
 \caption{Standard deviation of the mean for DBM results in the main paper}
 \label{tab:dbm_results}
 \smallskip
\centering
\resizebox{\textwidth}{!}{%
  \begin{tabular}{lcccccccccc}
  \toprule
  Method & \bfseries Adult & \bfseries Conn4 & \bfseries Digits & \bfseries DNA& \bfseries Mushr& \bfseries NIPS& \bfseries OCR& \bfseries RCV1& \bfseries Web \\
  \midrule
AdVIL-BP  & \num{2.56e-4}  & \num{1.81e-4}  & \num{1.03e-3}  & \num{7e-4}  & \num{5.89e-4}  & \num{2.98e-4}  & \num{4.05e-4}  & \num{1.02e-4}  & \num{1.01e-4} \\
AdVIL-Gibbs  & \num{1.27e-4} & \num{1.08e-4}  & \num{1.03e-3}  & \num{3.75e-4}  & \num{1.61e-3}  & \num{3.93e-4}  & \num{6.28e-4}  & \num{1.30e-4}  & \num{1.94e-4} \\
QT (Ours)  & \num{1.21e-3}  & \num{5.17e-4}  & \num{2.82e-4}  & \num{2.53e-4}  & \num{4.29e-4}  & \num{4.27e-4}  & \num{6.96e-4}  & \num{2.21e-4}  & \num{7.74e-5} \\
\bottomrule
  \end{tabular}}
\end{table*}

\subsection{Gaussian restricted Boltzmann machine (GRBM)}

\begin{table*}[h!]
\centering
\caption{Standard deviation of the mean for GRBM results in the main paper}
\label{tab:grbm_std_error}
\smallskip
  \begin{tabular}{lcc}
  \toprule
  Method & \bfseries Exp. 1 & \bfseries Exp. 2 \\
  \midrule
AdVIL-Gibbs  & \num{9.31e-5} & \num{1.08e-3}\\
AdVIL-BP  & \num{1.91e-5} & \num{2.58e-3} \\
QT (Ours)  &  \num{8.64e-4} & \num{1.06e-3} \\
\bottomrule
  \end{tabular}
\end{table*}

\subsection{Grid Markov random field (GMRF)}
\begin{table}[!htbp]
\centering
\caption{Standard deviation of the mean for GMRF results in the main paper. The \textsc{Random} approach is fully deterministic, so there is no difference in results between independent runs.}
\label{tab:GMRF}
\medskip
\begin{tabular}{lcc}
\toprule
Method & IOU & NCE \\
\midrule
\textsc{QT (Ours)} & $0.28$ & $0.34$ \\
\textsc{Random} & $9.0 \times 10^{-3}$ & $0$ \\
\bottomrule
\end{tabular}
\end{table}

\section{QT-NN architectures, derived for each PGM class}

Here we derive efficient and numerically robust forms of the BP update equations for each of our models. These form the QT-NNs used and trained in the main paper. %Although they follow applying  standard loopy BP to each PGM, they are provided here for completeness.

\subsection{Restricted Boltzmann machine (RBM)}
\label{sec:rbmcase}

\begin{figure*}[!htbp]
  \includegraphics[width=\linewidth]{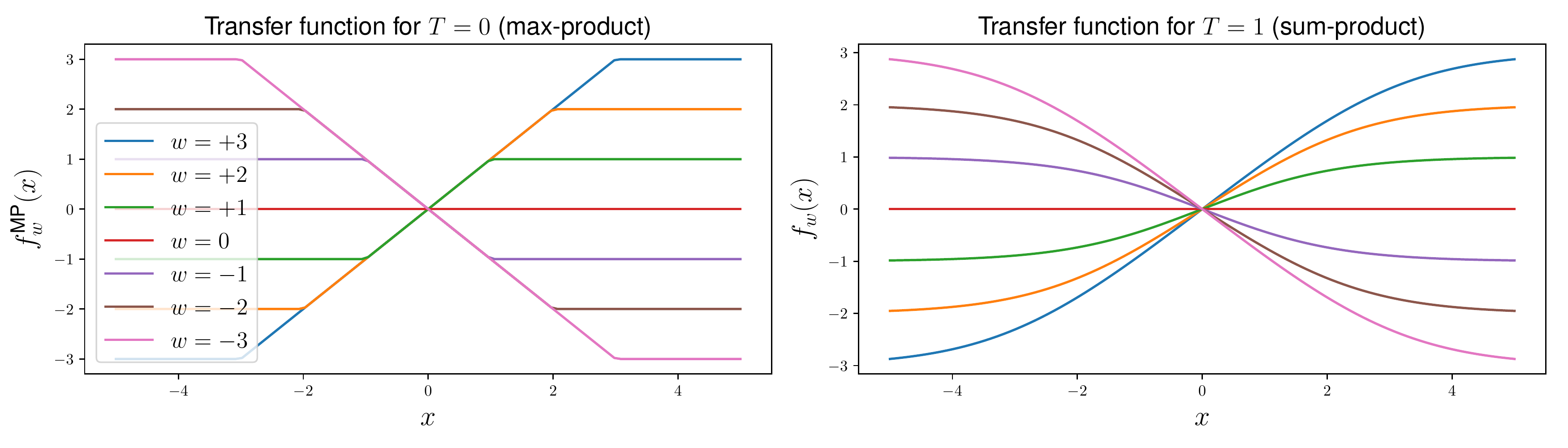}
  \caption{Transfer function for binary pairwise factors (with score 0 for agreement and $-w$ for disagreement) at two temperatures, max-product and sum-product. With this parameterization, setting a weight or input to zero results in zero output.}
  \label{fig:transfer}
\end{figure*}

\paragraph{RBM potential: }
We consider the case where the underlying PGM is a binary RBM with $H$ hidden units and $V$ visible units. We will use a slightly different parameterization (a linear transformation of the standard one) to simplify the form of the obtained transfer function. Thus, we set
\begin{equation}\label{eqn:rbm}
\phi({\bm v}, {\bm h}; \theta) = 2{\bm h}^\top W{\bm v} + {\bm h}^\top (\bm{c}_H-W\bm{1}_V) + {\bm v}^\top (\bm{c}_V-W^\top1_H)
\end{equation}
Note that the above potential can be expressed as:
\begin{align*}
\begin{split}
\phi({\bm v}, {\bm h}; \theta) &=
\frac{1}{2} (2{\bm h} - \bm{1}_H)^\top W (2 {\bm v} - \bm{1}_V) + \frac{1}{2} (2{\bm h} - \bm{1}_H)^\top \bm{c}_H +  \frac{1}{2} (2{\bm v} - \bm{1}_V)^\top \bm{c}_V \\
& ~~~~ - \frac{1}{2} \bm{1}_H^\top W\bm{1}_V +\frac{1}{2} \bm{c}_H^\top \bm{1}_H +\frac{1}{2} \bm{c}_V^\top \bm{1}_V.
\end{split}
\end{align*}
We can therefore define $\tilde{\bm h} = 2{\bm h} - \bm{1}_H~;~\tilde{\bm v} = 2{\bm v} - \bm{1}_V$ and consider the binary RBM where the hidden and visible variables take their values in $\{-1, 1\}$ and which has the potential:
$$
\tilde{\phi}(\tilde{\bm v}, \tilde{\bm h}; \theta) = \frac{1}{2} \tilde{\bm h}^\top W \tilde{\bm v} + \frac{1}{2} \tilde{\bm h}^\top {\bm c}_H + \frac{1}{2} \tilde{\bm v}^\top {\bm c}_V.
$$
$\phi$ and $\tilde{\phi}$ only differ by a constant of the model parameters, which will be cancelled out by the partition function. The RBM models are therefore equivalent. Three kinds of potentials are involved in $\tilde{\phi}$:
\newline
$\bullet$ potentials between hidden and visible variables: $\psi_{ij}(\tilde{h}_i, \tilde{v}_j) = \exp \left( \frac{1}{2} ~ w_{ij} ~ \tilde{h}_i ~ \tilde{v}_j \right), ~ \forall i,j$.
\newline
$\bullet$ potentials associated with hidden variables: $\kappa_{i}(\tilde{h}_i) = \exp \left( \frac{1}{2} ~ c_{H, i} ~ \tilde{h}_i \right), ~ \forall i$.
\newline
$\bullet$ potentials associated with visible variables: $\theta_{j}(\tilde{v}_j) = \exp \left( \frac{1}{2} ~ c_{V, j} ~ \tilde{v}_j \right), ~ \forall j$.

\paragraph{BP updates: }
We denote $N_{HV} \in \mathbb{R}_+^{H\times V \times 2}$ (resp. $N_{VH} \in \mathbb{R}_+^{V\times H \times 2}$) the messages going from the visible variables to the hidden variables (resp. from the hidden variables to the visible variables).

For an observed vector $\mathbf{x} \in \{-1, 1\}^V$ and a query vector $\mathbf{q} \in \{0, 1\}^V$, we note $\phi(v_i, x_i)$ the bottom-up messages to the $i$th visible variable. We have $\phi(-1, x_i)=\phi(1, x_i)=0.5$ if $x_i$ is being queried ($q_i = 0$); and $\phi(v_i, x_i) = 1$ if $v_i=x_i$ and $0$ otherwise if $x_i$ is being observed ($q_i=1$). It therefore holds: $\phi(-1, x_i) + \phi(1, x_i) =1$. All hidden variables are considered as being queried.

For the sake of simplicity, we first derive the BP updates for the sum-product BP algorithm with parallel schedules, before generalizing to a general temperature $T$. The message going from the $j$th visible variables to the $i$th hidden variables can be expressed as
\begin{equation}\label{BP-real-space}
n^{ij}_{HV}(\tilde{h}_j) ~~\propto
\sum_{\tilde{v}_j \in \{-1, 1\} } \psi_{ij}(\tilde{h}_i, \tilde{v}_j) \phi(\tilde{v}_j, x_j) \theta_j(\tilde{v}_j) \prod_{k \ne i} n^{jk}_{VH}(\tilde{v}_j), ~~ \forall \tilde{h}_i \in \{-1, 1\},
\end{equation}
which is equivalent to
\begin{align}\label{BP-log-space}
\begin{split}
n^{ij}_{HV}(\tilde{h}_i) ~~\propto ~~
& \psi_{ij}(\tilde{h}_i, -1) \phi(-1, x_j) \theta_j(-1) \prod_{k \ne i} n^{jk}_{VH}(-1) \\
&+ \psi_{ij}(\tilde{h}_j, 1) \phi(1, x_i) \theta_j(1) \prod_{k \ne i} n^{jk}_{VH}(1), ~~~ \forall \tilde{h}_i \in \{-1, 1\}
\end{split}
\end{align}
The BP updates for messages going from the hidden variables to the visible ones can be expressed in a similar fashion.

Numerical stability can be encouraged with the two following steps. First, we normalize messages: $$n^{ij}_{HV}(1) + n^{ij}_{HV}(-1) = 1.$$
Second, we map the messages to the logit space. Messages can then be expressed by a single float. We denote $M_{HV} \in \mathbb{R}^{H\times V}$ and $M_{VH} \in \mathbb{R}^{V\times H}$ the messages in the logit space. The sum-product BP updates in Equation \eqref {BP-log-space} are equivalent to:
\begin{align*}
m^{ij}_{HV} &= \text{logit}\left( n^{ij}_{HV}(1) \right) = \log\left( n^{ij}_{HV}(1) \right) - \log\left(1 - n^{ij}_{HV}(1) \right) \\
&= \log\left( n^{ij}_{HV}(1) \right) - \log\left(n^{ij}_{HV}(-1) \right) \\
&= \log \left( \psi_{ij}(1, -1) \phi(-1, x_j) \theta_j(-1) \prod_{k \ne i} n^{jk}_{VH}(-1)
+ \psi_{ij}(1, 1) \phi(1, x_j) \theta_j(1) \prod_{k \ne i} n^{jk}_{VH}(1) \right)\\
&~~~ - \log \left( \psi_{ij}(-1, -1) \phi(-1, x_j) \theta_j(-1) \prod_{k \ne i} n^{jk}_{VH}(-1)
+ \psi_{ij}(-1, 1) \phi(1, x_j) \theta_j(1) \prod_{k \ne i} n^{jk}_{VH}(1) \right)\\
&= \log \left( \psi_{ij}(1, -1) \phi(-1, x_j) \theta_j(-1)
+ \psi_{ji}(1, 1) \phi(1, x_j) \theta_j(1) \prod_{k \ne i} \exp \left(m^{jk}_{VH} \right)\right)\\
&~~~ - \log \left( \psi_{ij}(-1, -1) \phi(-1, x_j) \theta_j(-1) +
\psi_{ij}(-1, 1) \phi(1, x_j) \theta_j(1) \prod_{k \ne i} \exp \left(m^{jk}_{VH} \right) \right) \\
&= \log \left( 1
+ \frac{\psi_{ij}(1, 1) \phi(1, x_j) \theta_j(1)}
{\psi_{ij}(1, -1) \phi(-1, x_j) \theta_j(-1)}
\prod_{k \ne i} \exp \left(m^{jk}_{VH} \right)\right)
+ \log \left(\psi_{ji}(1, -1) \phi(-1, x_j) \theta_j(-1) \right)
\\
&~~~ - \log \left( 1
+ \frac{\psi_{ij}(-1, 1) \phi(1, x_j) \theta_j(1)}{\psi_{ij}(-1, -1) \phi(-1, x_j) \theta_j(-1)} \prod_{k \ne i} \exp \left(m^{jk}_{VH} \right)\right)
- \log \left(\psi_{ij}(-1, -1) \phi(-1, x_j) \theta_j(-1) \right)\\
&= \log \left( 1
+ \exp \left\{  w_{ij} + u_j + c_{V, j} + \sum_{k \ne i}  m^{jk}_{VH} \right\}\right)
 - \log \left( 1
+ \exp \left\{ - w_{ij} + u_j + c_{V, j} + \sum_{k \ne i}  m^{jk}_{VH} \right\}\right) \\
&~~~ - w_{ij}\\
&=f_{w_{ij}} \left( u_j + c_{V, j} + \sum_{k \ne i}  m^{jk}_{VH} \right),
\end{align*}
where we have defined $f_w(x) = \log(1 + e^{x + w}) - \log(1 + e^{x - w}) -w$ and $u_j = \text{logit} ~ \phi(1, x_j)$\footnote{In practice we clip the messages so that $u_i = -1000, ~0, ~1000$ for the values $\phi(1, x_j) = 0, ~0.5, ~1$.}. Note that we can simply express $u_j = \text{logit} \left( \frac{1 + x_j}{2} \right) q_j$.

\medskip
For numerical stability of the message updates, we use the property:
\begin{align*}
f_w(x) &= \log(1 + e^{x + w}) - \log(1 + e^{x - w}) -w
= \log(1 + e^{x + w}) - \log(e^w + e^{x})\\
&=
\operatorname{sign}(w) x|_{-|w|}^{|w|} + \log(1+e^{-|x + w|}) - \log(1+e^{-|x - w|}).
\end{align*}
After running the BP updates for $N$ iterations, we compute the beliefs (in the logit space) for the $j$th visible variable
\begin{align*}
b_{V, j}
&= \log \left( \phi(1, x_j) \theta_j(1) \prod_{k=1}^H n^{jk}_{VH}(1) \right) - \log \left( \phi(-1, x_j) \theta_j(-1) \prod_{k=1}^H n^{jk}_{VH}(-1) \right)\\
&= u_j + c_{V, j} + \sum_{k=1}^H  m^{jk}_{VH},
\end{align*}
and map the beliefs back to the real space by considering the sigmoid of this quantity.

\paragraph{BP summary:} We can summarize the architecture of the QT-NN for a general temperature $T$ by the following equations (which correspond to the unrollment of parallel BP over time using messages in logit space presented above):
\begin{align}
\bm{u}_V &= \operatorname{logit}\left( \frac{\bm{x} + 1}{2} \right)\circ\bm{q} &\nonumber\text{(unary term for visible units)}\\
\bm{u}_H &= \bm{0}_H &\nonumber\text{(unary term for hidden units)}\\
M_{HV}^{(0)} &= \bm{0}_{HV} &\nonumber\text{(init messages to 0)}\\
M_{VH}^{(0)} &= \bm{0}_{VH} &\nonumber\text{(init messages to 0)}\\
M_{HV}^{(n)} &= f_{W^\top}\left( \bm{u}_V \bm{1}_H^T + \bm{c}_V \bm{1}_H ^T + M_{VH}^{(n-1)} \bm{1}_{HH} - M_{VH}^{(n-1)} \right)^{\top}&\nonumber\text{(interlayer connection)}\\
M_{VH}^{(n)} &= f_{W} \left(\bm{u}_H \bm{1}_V^T + \bm{c}_H \bm{1}_V^T + M_{HV}^{(n-1)} \bm{1}_{VV}  - M_{HV}^{(n-1)} \right)^\top &\nonumber\text{(interlayer connection)}\\
\hat{\bm{v}} &= \sigma \left( \bm{u}_V + \bm{c}_V + M_{VH}^{(N)}\bm{1}_H \right) &\nonumber\text{(output layer for visible)}\\
\hat{\bm{h}} &= \sigma\left(\bm{u}_H + \bm{c}_H + M_{HV}^{(N)}\bm{1}_V \right) &\nonumber\text{(output layer for hidden)},
\end{align}
where
\begin{align}
\sigma(x) &= 1/(1+e^{-x}) & \nonumber \\
\operatorname{logit}(x) &= \sigma^{-1}(x) = \log(x) - \log(1-x) &\nonumber\\
%\end{align}\begin{align}
f_w^{\text{MP}}(x) &= \operatorname{sign}(w) x|_{-|w|}^{|w|}  &\nonumber\text{($a|_{b}^{c}$ truncates $a$ between $b$ and $c$, Fig.~\ref{fig:transfer} left)}\\
f_w(x) &= f_w^{\text{MP}}(x) + \operatorname{sp}(-|x + w|, T) \nonumber - \operatorname{sp}(-|x - w|, T) &\nonumber\text{(Fig.~\ref{fig:transfer} right) }\\
\operatorname{sp}(x, T) &= T \log(1+e^{x/T})&\nonumber\text{(a.k.a. softplus function)}.
\end{align}

We have used the following notations:
\begin{itemize}
  \item $\bm{0}_{HV}$ represents a matrix of zeros of size $H \times V$. Similarly $\bm{1}_{HH}$ represents a matrix of ones of size $H \times H$, and  $\bm{1}_{V}$ represents a matrix of ones of size $V \times 1$.
  \item When any of the above defined scalar functions is used with matrix arguments, the function is applied elementwise.
\end{itemize}

Some observations can be made:
\begin{itemize}
  \item The Hadamard product with $\bm{q}$ effectively removes the information from the elements of $\bm{v}$ not present in the query mask, replacing them with 0, which corresponds to a uniform binary distribution in logit space.
  \item The output of the network is $\hat{\bm{v}}$ and $\hat{\bm{h}}$, the inferred probability of 1 for both the visible and hidden units. The output $\hat{\bm{h}}$ is inferred but actually not used during training.
  \item The computation of $f_w(x)$ as specified above is designed to be numerically robust. It starts by computing $f_w^{\text{MP}}(x)$, which would be the value of $f_w(x)$ for a temperature $T=0$, i.e., max-product message passing, and then performs a correction on top for positive temperatures.
\end{itemize}

\subsection{Deep Boltzmann machine (DBM)}
\label{sec:dbmcase}
In this section we consider the slightly more complicated case in which the underlying PGM is a binary DBM with $V$ visible units and two hidden layers with $H_1$ and $H_2$ units. As in Section~\ref{sec:rbmcase}, we use the slightly different parametrization
\begin{align}\label{eqn:dbm}
\begin{split}
 &\phi({\bm v}, {\bm h}; \theta)\\
 = &2{\bm h_2}^\top W_{H_2 H_1}{\bm h_1} + 2{\bm h_1}^\top W_{H_1 V}{\bm v} + {\bm h_2}^\top (\bm{c}_{H_2}-W_{H_2 H_1}\bm{1}_{H_1})  + {\bm v}^\top (\bm{c}_V-W_{H_1 V}^\top1_{H_1})\\
 + & {\bm h_1}^\top (\bm{c}_{H_1}-W_{H_2 H_1}^\top\bm{1}_{H_2} - W_{H_1 V}\bm{1}_V)
\end{split}
\end{align}
We note that, due to the fact that we are considering a DBM with just two hidden layers, we can reuse the equations derived in Section~\ref{sec:rbmcase} by some simple transformations. More concretely, define
$$\bm{\tilde{h}} = \bm{h_1}, \bm{\tilde{v}} = \begin{bmatrix}\bm{v}\\ \bm{h_2}\end{bmatrix}, \tilde{W} = \begin{bmatrix}W_{H_1 V}\\ W_{H_2 H_1}^\top\end{bmatrix}, \bm{\tilde{c}}_H = \bm{c}_{H_1}, \bm{\tilde{c}}_{V} = \begin{bmatrix}\bm{c}_V\\ \bm{c}_{H_2}\end{bmatrix}$$
It's easy to see that plugging $\tilde{\bm{h}}, \tilde{\bm{v}}, \tilde{W}, \tilde{\bm{c}}_H, \tilde{\bm{c}}_V$ into Equation~\ref{eqn:rbm} recovers Equation~\ref{eqn:dbm}.

As a result, the QT-NN equations derived in Section~\ref{sec:rbmcase} for the RBM apply equally well to the DBM with two hidden layers. The only change needed lies in the incoming messages and the queries. Use $v, q$ to denote the original observed values for the visible units and our interested query. The corresponding unary term for visible units for DBM is define as $\bm{u}_V = \operatorname{logit}(\bm{\tilde{v}})\circ\bm{\tilde{q}}$, where
$$\bm{\tilde{v}} = \begin{bmatrix}\bm{v} \\ 0.5\bm{1}_{H_2}\end{bmatrix}\text{ and }\bm{\tilde{q}} = \begin{bmatrix}\bm{q} \\ \bm{1}_{H_2}\end{bmatrix}$$

\subsection{Gaussian restricted Boltzmann machine (GRBM)}
\label{sec:grbmcase}
The log-probability of a GRBM is proportional to:
\begin{equation}
\phi({\bm v}, {\bm h}; \theta) = -\frac{1}{2 \sigma^2} \|{\bm v}- {\bm b}\|^2 + {\bm c}^\top {\bm h} + \frac{1}{\sigma} {\bm h}^\top W {\bm v}
\end{equation}
where $\sigma$ is fixed to $1$. To embed passing of continuous messages in the QT-NN, we approximate continuous messages with Gaussian distributions. In the following equations, for every message to a visible unit, we store the two natural parameters of a Gaussian distribution corresponding to that message.

\begin{align}
\bm{u}_H &= \bm{0}_H &\nonumber\text{(unary term for hidden units)}\\
\bm{u}_{V\theta_1} &= \bm{v} \circ \bm{q} + \bm{b} \circ(\bm{1}_V - \bm{q}) &\nonumber\text{(param 1 for visible unary term)}\\
\bm{u}_{V\theta_2} &= -\frac{1}{2\epsilon} \cdot \bm{q} -\frac{1}{2} \cdot  (\bm{1}_V - \bm{q}) &\nonumber\text{(param 2 of visible unary term)}\\
M_{HV}^{(0)} &= \bm{0}_{HV} &\nonumber\text{(init message from visible to hidden)}\\
M_{VH\theta_1}^{(0)} &= \bm{0}_{VH} &\nonumber\text{(init message from hidden to visible)}\\
M_{VH\theta_2}^{(0)} &= -\frac{1}{2} \cdot \bm{1}_{VH} &\nonumber\text{(init message from hidden to visible)}\\
C_{VH\theta_1}^{(n-1)} & =\bm{u}_{V\theta_1} + M_{VH\theta_1}^{(n-1)} \bm{1}_H - M_{VH\theta_1}^{(n-1)} &\nonumber\text{(param 1 of cavity)}\\
C_{VH\theta_2}^{(n-1)} & =\bm{u}_{V\theta_2} + M_{VH\theta_2}^{(n-1)} \bm{1}_H - M_{VH\theta_2}^{(n-1)} &\nonumber\text{(param 2 of cavity)}\\
M_{HV}^{(n)} &=  f_{W^\top} (C_{VH\theta_1}^{(n-1)}, C_{VH\theta_2}^{(n-1)})^\top  &\nonumber\text{(interlayer connection)}\\
C_{HV}^{(n-1)} &= \bm{u}_{H} + \bm{c}_H + M_{HV}^{(n-1)} \bm{1}_V - M_{HV}^{(n-1)} &\nonumber\text{(cavity)}\\
M_{VH\theta_1}^{(n)},M_{VH\theta_2}^{(n)}  &=  g_{W} (C_{HV}^{(n-1)}, C_{VH\theta_1}^{(n-1)}, C_{VH\theta_2}^{(n-1)}) &\nonumber\text{(interlayer connection)}\\
\hat{\bm{v}}_{\theta_1} &= \bm{u}_{V\theta_1} + M_{VH\theta_1}^{(N)} \bm{1}_H &\nonumber\text{(output layer for visible)}\\
\hat{\bm{v}}_{\theta_2} &= \bm{u}_{V\theta_2} + M_{VH\theta_2}^{(N)} \bm{1}_H &\nonumber\text{(output layer for visible)}\\
\hat{\bm{h}} &= \sigma(\bm{u}_H + \bm{c}_H + M_{HV}^{(N)}\bm{1}_V) &\nonumber\text{(output layer for hidden)},
\end{align}
where
\begin{align}
f_w (x, y) &= - \frac{2xw + w^2}{4y} &\nonumber \\
g_w (x, \theta_1, \theta_2) &= (b_1(w, x, \theta_1, \theta_2) - \theta_1, b_2(w, x, \theta_1, \theta_2) - \theta_2)& \nonumber\\
\sigma(x) &= 1/(1+e^{-x}) & \nonumber
\end{align}

Notation clarifications:
\begin{itemize}
  \item $\theta_1$ and $\theta_2$ are used to denote the two natural parameters of a Gaussian distribution. In the case of $M_{VH\theta_1}$ and $M_{VH\theta_2}$, these are parameters approximating the messages from hidden units to visible units with a Gaussian distribution..
  \item The functions $b_1$ and $b_2$ output the two natural parameters of a Gaussian distribution approximating the belief at a visible unit, as is described below.
\end{itemize}

Given $w$ (the weight of the connection between a hidden unit and visible unit), $x$ (the cavity at the hidden unit), and $\theta_1$, $\theta_2$ (the natural parameters of the cavity at the visible unit), we approximate the belief at the visible unit as follows:

\begin{align}
\mu &=  \frac{-\theta_1}{2 \theta_2}\nonumber \\
\sigma^2 &=  \frac{-1}{2 \theta_2}\nonumber \\
\mu_B &= \frac{e^{x+w\mu + \frac{1}{2} \sigma^2 w^2} (\mu + \sigma^2 w) + \mu}{e^{x+w\mu + \frac{1}{2} \sigma^2 w^2} + 1}\nonumber\\
\sigma^2_B &= \frac{e^{x+w\mu + \frac{1}{2} \sigma^2 w^2} ((\mu + \sigma^2 w)^2 + \sigma^2) + \mu^2 + \sigma^2}{e^{x+w\mu + \frac{1}{2} \sigma^2 w^2} + 1} - \mu_B^2 \nonumber\\
\end{align}

The approximation of the belief is a Gaussian with mean $\mu_B$ and variance $\sigma^2_B$. and the functions $b_1$ and $b_2$ return the two natural parameters of that Gaussian,  $\frac{\mu_B}{\sigma^2_B}$ and $-\frac{1}{\sigma^2_B}$ respectively.

\bibliography{query_training}
\end{appendices}